\documentclass[11pt,a4paper]{article}
\usepackage[hyperref]{eacl2021}
\usepackage{times}
\usepackage{latexsym}
\usepackage{lipsum}
\usepackage{adjustbox}
\usepackage{tablefootnote}

\usepackage{microtype}
\usepackage{multirow}
\usepackage{float}
\usepackage{dblfloatfix}
\aclfinalcopy

\title{Hierarchical Transformer Model for Scientific Named Entity Recognition}

\author{Urchade Zaratiana$^{*\dagger}$, Pierre Holat$^{*\dagger}$, Nadi Tomeh$^\dagger$, Thierry Charnois$^{\dagger}$ \\
$^\star$ FI Group, Puteaux, France \\
$^\dagger$ LIPN, Université Sorbonne Paris Nord - CNRS UMR 7030, Villetaneuse, France\\
  {\tt \{urchade.zaratiana,pierre.holah\}@fi-group.com} \\
  {\tt \{charnois,tomeh\}@lipn.fr} \\}

\date{}

\begin{document}
\maketitle
\begin{abstract}
The task of Named Entity Recognition (NER) is an important component of many natural language processing systems, such as relation extraction and knowledge graph construction. In this work, we present a simple and effective approach for Named Entity Recognition. The main idea of our approach is to encode the input subword sequence with a pre-trained transformer such as BERT, and then, instead of directly classifying the word labels, another layer of transformer is added to the subword representation to better encode the word-level interaction. We evaluate our approach on three benchmark datasets for scientific NER, particularly in the computer science and biomedical domains.  Experimental results show that our model outperforms the current state-of-the-art on SciERC and TDM datasets without requiring external resources or specific data augmentation. Code is available at \url{https://github.com/urchade/HNER}.
\end{abstract}

\section{Introduction}
\paragraph{} Extracting entities from scientific texts is an important task of Natural Language Processing. While traditional methods were based on manually generated features, pre-trained language models such as BERT \citep{devlin-etal-2019-bert} have recently achieved competitive results. However, since BERT was pre-trained on general text from Wikipedia and Bookcorpus, its performance on domain-specific tasks has been shown to be suboptimal in several previous works \citep{beltagy-etal-2019-scibert,Lee_2019}. Those empirical findings have motivated the development of domain-specific pre-trained language models. In particular, several domain-specific transformers have been made publicly available through the hugging face's transformer library. For example, there are SciBERT \citep{beltagy-etal-2019-scibert} for the scientific domain, BioBERT \citep{Lee_2019} for the biomedical domain, and FinBERT \citep{yang2020finbert} for the financial domain.
\begin{figure}
    \centering
    \includegraphics[width=1.\columnwidth]{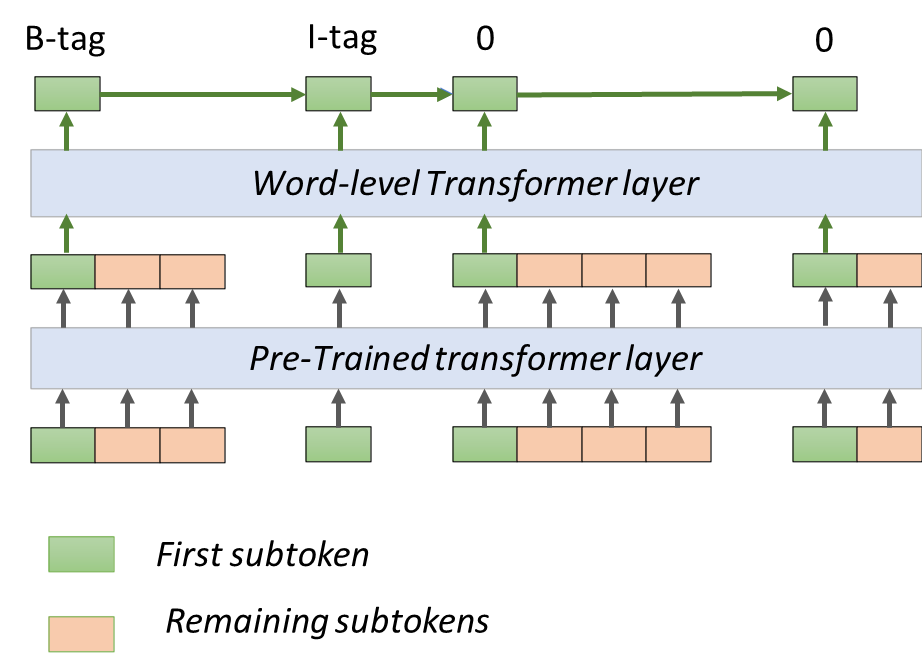}
    \caption{This figure shows the main architecture of our model. After tokenization, each subwords are passed into a pre-trained model. Then, another transformation layer represents each word using its first subword representation prior to being classified.}
    \label{fig:my_label}
\end{figure}
\paragraph{} Pre-trained transformers such as SciBERT or BioBERT have achieved excellent performance in scientific NER compared to previous work. However, despite this success, their traditional fine-tuning for NER can be suboptimal, as they typically classify the first subtoken representation of each word to label sequences. Some work has attempted to avoid this problem by designing NER as a span-based classification instead of sequence labeling \citep{Eberts2020SpanbasedJE}. However, these methods are more difficult to implement and require many additional hyperparameters such as the number of negative samples and window size. In this work, we adopt the conventional sequence labeling approach using the BIO scheme, but instead of directly classifying word labels, a transformer layer \citep{vaswani2017attention} is added on top of the subword representation to better encode the word-level interaction. More specifically, each first subtoken of every word is passed through an additional transformer layer before the word classification. This small change in the architecture can significantly induce additional performance.

\section{Model}

\paragraph{} In this paper, we treat the NER task as a classification of BIO tags. Our model consists of three main elements: a pre-trained transformer layer, a word-level interaction layer and a classification layer composed of a linear layer and a CRF layer.

\paragraph{Pre-trained transformer layer} The first part of our architecture is a pre-trained transformer model to encode input subwords into contextualized embeddings.  Usually, a model like BERT \citep{devlin-etal-2019-bert} or RoBERTa \citep{liu2019roberta} is considered for this step, but since we deal with scientific documents, we used a domain-specific model called Scibert \citep{beltagy-etal-2019-scibert}.

\paragraph{Word-level layer} This component takes as input the first subword of each word in the previous layer and encodes their interaction with a single-layer transformer \citep{vaswani2017attention}.  Unlike previous models that directly classify the first subtoken, the addition of this layer provides a better representation for the sequence labeling.

\paragraph{Classification layer} Finally, this component classifies each word representation into the corresponding label using a linear layer. Furthermore, with the use of a CRF layer \citep{10.5555/645530.655813}, we ensure that all output labels are valid by adding constraints during inference (e.g., label O must not precede label I) whitch also make evaluation easier.

\section{Experiments}
\subsection{Datasets}

\paragraph{} SciERC is a dataset introduced by \citep{luan-etal-2018-multi} for scientific information extraction. The dataset consists of 500 abstracts extracted from papers related to Artificial Intelligence. They have been annotated with scientific entities, their relations and conference clusters. 

\paragraph{} TDM is a NER dataset that was recently published \citep{hou2021tdmsci} for detecting Tasks, Datasets, and Metrics (TDM) from Natural Language Processing papers. It consists of sentences extracted from the full text of natural language processing papers, not just abstracts.

\paragraph{} NCBI \citep{DOGAN20141} is a NER dataset that is designed to identify disease mentions in biomedical texts from PubMed article abstracts.

\begin{table}[h!]
\centering
 \begin{tabular}{c | c | c | c | c} 
  & Domain & Train & Dev & Test \\
\hline\hline
 TDM & CS & 1523 &  & 487 \\ 

 NCBI & Bio & 5432 & 923 & 940 \\ 

 SciERC & CS & 350 & 50 & 50 \\ 

 \end{tabular}
 \caption{\label{table_1} The statistics of the datasets}
\end{table}

\begin{table*}
\centering
\begin{adjustbox}{width=0.7\textwidth}
 \begin{tabular}{c | c | c| c | c} 
Dataset  & Model & Precision & Recall & F1 \\
\hline
\hline
\multirow{8}{*}{SciERC}  
                & \citet{beltagy-etal-2019-scibert} &  &  & 67.57$^{\ddagger}$ \\
                        & \citet{zhong-chen-2021-frustratingly} &  &  & 68.90$^\dagger$ \\
                        & \citet{Eberts2020SpanbasedJE} & 70.87$^\dagger$ & 69.79$^\dagger$ & 70.33$^\dagger$ \\
                        & \multirow{2}{*}{Ours (LSTM)}
                        & 67.03$^\dagger$ &    75.22$^\dagger$  &   70.89$^\dagger$ \\
                        && 66.29$^{\ddagger}$ &    72.67$^{\ddagger}$  &   69.19$^{\ddagger}$ \\
                         & \multirow{2}{*}{Ours (Transformer)} 
                         & 67.99$^\dagger$ &    74.11$^\dagger$  &  \textbf{70.91}$^\dagger$ \\
                        & & 67.50$^{\ddagger}$ &    72.22$^{\ddagger}$  &  \textbf{69.64}$^{\ddagger}$ \\
\hline\hline                        
\multirow{3}{*}{TDM}  
                        & \citet{hou2021tdmsci}$^\dagger$ & 67.17 & 58.27 & 62.40 \\

                        & Ours (LSTM)$^\dagger$ & 64.79  &  71.00   & 67.75 \\
                        & Ours (Transformer)$^\dagger$ 
                        & 65.56 &   70.21 &  \textbf{67.81} \\
\hline\hline                        
\multirow{4}{*}{NCBI-disease}  & \citet{beltagy-etal-2019-scibert}$^\dagger$ &  &  & 88.57 \\
                        & \citet{Lee_2019}$^\dagger$ & 88.22 & 91.25 & \textbf{89.71}  \\
                        & Ours (LSTM)$^\dagger$ & 84.66 & 89.69 &   87.10 \\
                        & Ours (Transformer)$^\dagger$ & 86.33 &    90.10  &  88.18 \\

 \end{tabular}
 \end{adjustbox}
 \caption{\label{table_1}
This table compares the performance of our model to the state of the art on named entity recognition for scientific texts. The results presented here are the best obtained from three different random seeds.  $^\dagger$: micro-average; $^{\ddagger}$: macro-average
}
\end{table*}

\subsection{Implementation details}

\paragraph{Architecture} For all experiments, we used the recommended version of SciBERT \citep{beltagy-etal-2019-scibert}, which contains 12 transform layers and a hidden dimension of 768 for the pre-trained model. In the case of the word-level layer, we employ a one-layer bidirectional transformer with a hidden dimension of 768 and 8 attention heads. We also experiment with a Bi-LSTM \citep{10.1162/neco.1997.9.8.1735} with a hidden dim of 768 in place of the word-level transformer layer to see the difference in performance. The final layer is a linear layer that is used to project the word representation into label space before feeding it into a CRF layer for decoding.

\paragraph{Hyperparameters} We did not conduct an extensive search for hyperparameters, but rather the ones recommended by \citep{devlin-etal-2019-bert}. For all datasets, we picked a learning rate of 3e-5 and used a batch size of 4 for TDM and SciERC and a batch size of 16 for the NCBI dataset. We trained all our models for up to 25 epochs and upon completion, we selected the checkpoints with the best f1 score on the validation set. 

Inspired by research on semi- and self-supervised learning \citep{tarvainen2018mean, he2020momentum}, we maintain an exponential moving average (EMA) of all parameters during learning. We find that this simple technique yields additional performance at almost no cost. More formally, after each gradient, $\theta_{k }^{'} = \lambda \theta_{k-1}^{'} + (1 - \lambda) \theta_{k} $ where $\theta_{k}$ and $\theta_{k}^{'}$ represent the parameters of the model and the parameter of the moving average at the step k, respectively. Furthermore, $\lambda$ is a hyperparameter for the EMA update that is usually set to a floating number close to 1.0; in our experiment, we used a $\lambda$ of 0.99.

All of our models were trained with a single V100 GPU.

\subsection{Evaluation}

\paragraph{} We evaluate the models on the exact correspondence between the gold entities and the predicted entities. Furthermore, in line with previous research, we exclude non-entity spans. We therefore consider precision, recall and F-score as metrics, implemented by the seqeval library \citep{seqeval}. For SciERC, we reported both micro and macro averages, however, we only reported the micro average for TDM and NCBI-Disease for comparison with previous works..

\subsection{Main results}

\paragraph{} The main results of our experimentation are presented in Table \ref{table_1}. It shows that our best models are able to outperform the state of the art on SciERC and TDM and obtain a competitive result on the NCBI dataset.

On SciERC, our model is able to surpass SciBERT \citep{beltagy-etal-2019-scibert} by a significant margin (+2 on F1-macro). Our method is also capable of outperforming span-based approaches such as PURE \citep{zhong-chen-2021-frustratingly} and Spert \citep{Eberts2020SpanbasedJE} while employing a much simpler approach. More specifically, we achieved 70.91 on F1-micro while SPERT, the current state of the art got 70.33.

Our model also exceeds the state of the art on the TDM dataset by 5 points on F1-micro without using any data augmentation technique.

Finally, our approach also showed competitive results on the NCBI dataset, while not exceeding the state of the art. Our model has roughly the same performance as the original SciBERT NER model. However, the current state of the art, the BioBERT NER model, outperforms our model by more than one point on the F1-micro

\subsection{Bi-LSTM vs Transformer}

\paragraph{} Table \ref{table_1} also reports our experimentation using a Bi-LSTM as the word-level layer. We can see that the transformer is more efficient than the Bi-LSTM layer for all datasets. However, the additional gain of the transformer may be due to the fact that it contains more parameters.

\subsection{Ablation studies}

\paragraph{} Here, we undertake an ablation study to see the contribution of different components of our proposed model. In detail, we examine the addition of the word-level layer and the effectiveness of the exponential moving average model. The reported results is the micro-F1 averaged accross three different random seeds.

\paragraph{Word level layer} In this study, we examine the effectiveness of our word-level interaction approach by comparing it to subword-level interaction with the same number of parameters. More concretely, the subword-level interaction is achieved by simply adding another layer of transformers on top of the scibert representation. The following table shows the comparison between the two approaches:

\begin{table}[H]
\centering
 \begin{tabular}{c | c | c} 
  & Word level & Subword level \\
\hline
\hline
 TDM & $66.69 \pm 1.17$ & $66.16 \pm 0.77$ \\ 

 NCBI & $87.81 \pm 0.33$ & $87.65 \pm 0.38$ \\ 

 SciERC & $70.58 \pm 0.34$ & $70.48 \pm 0.26$ \\ 
 \end{tabular}
 
  \caption{\label{table_2}
Ablation study: word-level vs subword level representation
}

\end{table}

As shown in the table \ref{table_2}, encoding the information at the word level is in fact beneficial. We demonstrate from the results of this experiment that the additional performance provided by our approach is not only due to the larger number of parameters but to the fact that it produces a better representation between words to predict NER tags. Intuitively, our approach forces each first subtoken to encode the entire word information through a self-attention mechanism, and then the addition of the transformation layer models their interaction.

\paragraph{Exponential Moving Average} We also investigate the effectiveness of keeping an Exponential Moving Average (EMA) of the parameters during training which consist in keeping a moving average of the model parameter after each gradient step.

\begin{table}[H]
\centering
 \begin{tabular}{c | c | c} 
  & Base Parameters & EMA Parameters \\
\hline
\hline
 TDM & $66.52 \pm 0.45$ & $66.69 \pm 1.17$ \\ 

 NCBI & $86.06 \pm 0.60$ & $87.81 \pm 0.33$ \\ 

 SciERC & $69.78 \pm 0.45$ & $70.58 \pm 0.34$ \\ 

 \end{tabular}
 \caption{\label{table_3} This table compare the performance between the base parameter and the ema parameter.}
\end{table}

This table \ref{table_3} shows that keeping an exponential moving average of the parameter during training can provide additional performance gain almost for free. The EMA version outperforms the baseline for all datasets on which we trained our model. In fact, keeping an moving average of the parameters can be seen as a soft ensembling of all the previous model checkpoints. Furthermore, keeping an EMA of model parameters could be employed for any deep learning task and not limited to Named Entity Recognition. It could be interesting to investigate its effectiveness across a wider range of NLP task, but it is not in the scope of our study so we leave it for future work.

\section{Related Work}

Early works in NER were made up of handcrafted features and CRF-like models. The advent of deep learning allowed for end-to-end model training. In the early days of deep learning, the majority of NER models used an LSTM-based architecture with word- and character-level encoding and CRF layer for decoding \citep{lample-etal-2016-neural, huang2015bidirectional}. Recently, the arrival of BERT has dramatically transformed the field of NLP. In terms of NER, BERT's paper proposed to do NER by classifying the hidden representation of the first subtoken of each word for sequence labeling. Then, this approach has been adapted to different domains such as the scientific domain through the works of \citet{beltagy-etal-2019-scibert} and \citet{Lee_2019} among others. For example, \citet{beltagy-etal-2019-scibert} use scibert as an encoder and project the first subtoken into the label space and then employ a CRF layer for decoding \citep{Eberts2020SpanbasedJE, zhong-chen-2021-frustratingly}. Some work has employed span-based approach as opposed to traditional sequence labelling. In the span-based approach, each span in a sequence up to a maximum sequence length is enumerated, aggregated, and then classified. These methods are particularly useful for modelling joint entity and relation extraction \citep{Wadden2019EntityRA} and nested named entity recognition.

\section{Conclusion}
In this paper, we proposed a new architecture for applied named entity recognition. Our model outperforms the state of the art on two scientific NER benchmarks, namely the SciERC and TDM datasets, and achieves competitive performance on the NCBI-disease dataset. The advantage of our model is that it is simple, does not require additional data or external resources such as gazetter to achieve competitive results. Our empirical results also show that maintaining an exponential moving average of the model parameters during learning can provide an additional performance gain for a negligible computational resource (time complexity).

\bibliography{eacl2021}

\begin{thebibliography}{19}
\expandafter\ifx\csname natexlab\endcsname\relax\def\natexlab#1{#1}\fi

\bibitem[{Beltagy et~al.(2019)Beltagy, Lo, and
  Cohan}]{beltagy-etal-2019-scibert}
Iz~Beltagy, Kyle Lo, and Arman Cohan. 2019.
\newblock \href {https://doi.org/10.18653/v1/D19-1371} {{S}ci{BERT}: A
  pretrained language model for scientific text}.
\newblock In \emph{Proceedings of the 2019 Conference on Empirical Methods in
  Natural Language Processing and the 9th International Joint Conference on
  Natural Language Processing (EMNLP-IJCNLP)}, pages 3615--3620, Hong Kong,
  China. Association for Computational Linguistics.

\bibitem[{Devlin et~al.(2019)Devlin, Chang, Lee, and
  Toutanova}]{devlin-etal-2019-bert}
Jacob Devlin, Ming-Wei Chang, Kenton Lee, and Kristina Toutanova. 2019.
\newblock \href {https://doi.org/10.18653/v1/N19-1423} {{BERT}: Pre-training of
  deep bidirectional transformers for language understanding}.
\newblock In \emph{Proceedings of the 2019 Conference of the North {A}merican
  Chapter of the Association for Computational Linguistics: Human Language
  Technologies, Volume 1 (Long and Short Papers)}, pages 4171--4186,
  Minneapolis, Minnesota. Association for Computational Linguistics.

\bibitem[{Doğan et~al.(2014)Doğan, Leaman, and Lu}]{DOGAN20141}
Rezarta~Islamaj Doğan, Robert Leaman, and Zhiyong Lu. 2014.
\newblock \href {https://doi.org/https://doi.org/10.1016/j.jbi.2013.12.006}
  {Ncbi disease corpus: A resource for disease name recognition and concept
  normalization}.
\newblock \emph{Journal of Biomedical Informatics}, 47:1--10.

\bibitem[{Eberts and Ulges(2020)}]{Eberts2020SpanbasedJE}
Markus Eberts and A.~Ulges. 2020.
\newblock Span-based joint entity and relation extraction with transformer
  pre-training.
\newblock \emph{ArXiv}, abs/1909.07755.

\bibitem[{He et~al.(2020)He, Fan, Wu, Xie, and Girshick}]{he2020momentum}
Kaiming He, Haoqi Fan, Yuxin Wu, Saining Xie, and Ross Girshick. 2020.
\newblock \href {http://arxiv.org/abs/1911.05722} {Momentum contrast for
  unsupervised visual representation learning}.

\bibitem[{Hochreiter and Schmidhuber(1997)}]{10.1162/neco.1997.9.8.1735}
Sepp Hochreiter and Jürgen Schmidhuber. 1997.
\newblock \href {https://doi.org/10.1162/neco.1997.9.8.1735} {{Long Short-Term
  Memory}}.
\newblock \emph{Neural Computation}, 9(8):1735--1780.

\bibitem[{Hou et~al.(2021)Hou, Jochim, Gleize, Bonin, and
  Ganguly}]{hou2021tdmsci}
Yufang Hou, Charles Jochim, Martin Gleize, Francesca Bonin, and Debasis
  Ganguly. 2021.
\newblock \href {http://arxiv.org/abs/2101.10273} {Tdmsci: A specialized corpus
  for scientific literature entity tagging of tasks datasets and metrics}.

\bibitem[{Huang et~al.(2015)Huang, Xu, and Yu}]{huang2015bidirectional}
Zhiheng Huang, Wei Xu, and Kai Yu. 2015.
\newblock \href {http://arxiv.org/abs/1508.01991} {Bidirectional lstm-crf
  models for sequence tagging}.

\bibitem[{Lafferty et~al.(2001)Lafferty, McCallum, and
  Pereira}]{10.5555/645530.655813}
John~D. Lafferty, Andrew McCallum, and Fernando C.~N. Pereira. 2001.
\newblock Conditional random fields: Probabilistic models for segmenting and
  labeling sequence data.
\newblock In \emph{Proceedings of the Eighteenth International Conference on
  Machine Learning}, ICML '01, page 282–289, San Francisco, CA, USA. Morgan
  Kaufmann Publishers Inc.

\bibitem[{Lample et~al.(2016)Lample, Ballesteros, Subramanian, Kawakami, and
  Dyer}]{lample-etal-2016-neural}
Guillaume Lample, Miguel Ballesteros, Sandeep Subramanian, Kazuya Kawakami, and
  Chris Dyer. 2016.
\newblock \href {https://doi.org/10.18653/v1/N16-1030} {Neural architectures
  for named entity recognition}.
\newblock In \emph{Proceedings of the 2016 Conference of the North {A}merican
  Chapter of the Association for Computational Linguistics: Human Language
  Technologies}, pages 260--270, San Diego, California. Association for
  Computational Linguistics.

\bibitem[{Lee et~al.(2019)Lee, Yoon, Kim, Kim, Kim, So, and Kang}]{Lee_2019}
Jinhyuk Lee, Wonjin Yoon, Sungdong Kim, Donghyeon Kim, Sunkyu Kim, Chan~Ho So,
  and Jaewoo Kang. 2019.
\newblock \href {https://doi.org/10.1093/bioinformatics/btz682} {Biobert: a
  pre-trained biomedical language representation model for biomedical text
  mining}.
\newblock \emph{Bioinformatics}.

\bibitem[{Liu et~al.(2019)Liu, Ott, Goyal, Du, Joshi, Chen, Levy, Lewis,
  Zettlemoyer, and Stoyanov}]{liu2019roberta}
Yinhan Liu, Myle Ott, Naman Goyal, Jingfei Du, Mandar Joshi, Danqi Chen, Omer
  Levy, Mike Lewis, Luke Zettlemoyer, and Veselin Stoyanov. 2019.
\newblock \href {http://arxiv.org/abs/1907.11692} {Roberta: A robustly
  optimized bert pretraining approach}.

\bibitem[{Luan et~al.(2018)Luan, He, Ostendorf, and
  Hajishirzi}]{luan-etal-2018-multi}
Yi~Luan, Luheng He, Mari Ostendorf, and Hannaneh Hajishirzi. 2018.
\newblock \href {https://doi.org/10.18653/v1/D18-1360} {Multi-task
  identification of entities, relations, and coreference for scientific
  knowledge graph construction}.
\newblock In \emph{Proceedings of the 2018 Conference on Empirical Methods in
  Natural Language Processing}, pages 3219--3232, Brussels, Belgium.
  Association for Computational Linguistics.

\bibitem[{Nakayama(2018)}]{seqeval}
Hiroki Nakayama. 2018.
\newblock \href {https://github.com/chakki-works/seqeval} {{seqeval}: A python
  framework for sequence labeling evaluation}.
\newblock Software available from https://github.com/chakki-works/seqeval.

\bibitem[{Tarvainen and Valpola(2018)}]{tarvainen2018mean}
Antti Tarvainen and Harri Valpola. 2018.
\newblock \href {http://arxiv.org/abs/1703.01780} {Mean teachers are better
  role models: Weight-averaged consistency targets improve semi-supervised deep
  learning results}.

\bibitem[{Vaswani et~al.(2017)Vaswani, Shazeer, Parmar, Uszkoreit, Jones,
  Gomez, Kaiser, and Polosukhin}]{vaswani2017attention}
Ashish Vaswani, Noam Shazeer, Niki Parmar, Jakob Uszkoreit, Llion Jones,
  Aidan~N. Gomez, Lukasz Kaiser, and Illia Polosukhin. 2017.
\newblock \href {http://arxiv.org/abs/1706.03762} {Attention is all you need}.

\bibitem[{Wadden et~al.(2019)Wadden, Wennberg, Luan, and
  Hajishirzi}]{Wadden2019EntityRA}
David Wadden, Ulme Wennberg, Yi~Luan, and Hannaneh Hajishirzi. 2019.
\newblock Entity, relation, and event extraction with contextualized span
  representations.
\newblock \emph{ArXiv}, abs/1909.03546.

\bibitem[{Yang et~al.(2020)Yang, UY, and Huang}]{yang2020finbert}
Yi~Yang, Mark Christopher~Siy UY, and Allen Huang. 2020.
\newblock \href {http://arxiv.org/abs/2006.08097} {Finbert: A pretrained
  language model for financial communications}.

\bibitem[{Zhong and Chen(2021)}]{zhong-chen-2021-frustratingly}
Zexuan Zhong and Danqi Chen. 2021.
\newblock \href {https://doi.org/10.18653/v1/2021.naacl-main.5} {A
  frustratingly easy approach for entity and relation extraction}.
\newblock In \emph{Proceedings of the 2021 Conference of the North American
  Chapter of the Association for Computational Linguistics: Human Language
  Technologies}, pages 50--61, Online. Association for Computational
  Linguistics.

\end{thebibliography}
\bibliographystyle{acl_natbib}

\end{document}